\documentclass[10pt, a4paper]{article}
\usepackage{lrec}
\usepackage{multibib}
\usepackage[latin1]{inputenc}
\usepackage [english]{babel}
\usepackage [autostyle, english = american]{csquotes}
\MakeOuterQuote{"}
\newcites{languageresource}{Language Resources}
\usepackage{color}
\usepackage{colortbl}
\usepackage{graphicx}
\usepackage{tabularx}
\usepackage{soul}
\usepackage{epstopdf}
\usepackage{url}
\usepackage{pgfplots}
\usepackage{booktabs}

\newcolumntype{P}[1]{>{\centering\arraybackslash}p{#1}}
\usepackage{tikz}
\usepackage{multirow}
\usepackage{hyperref}
\usepackage{xstring}
\usepackage{graphicx}
\usepackage{lineno,hyperref}
\usepackage{multirow}
\usepackage{listings,xcolor}
\usepackage{rotating}
\usepackage{listings}
\usepackage{setspace}
\usepackage{color, colortbl, xcolor}
\RequirePackage{fix-cm}
\definecolor{Gray}{rgb}{0.7,0.75,0.71}
\definecolor{aurometalsaurus}{rgb}{0.43, 0.5, 0.5}

\title{Exploring Conversational Language Generation \\ for Rich Content about Hotels}
\name{Marilyn A. Walker$^1$, Albry Smither$^2$, Shereen Oraby$^1$, Vrindavan Harrison$^1$, Hadar Shemtov$^3$}
\address{University of California Santa Cruz$^1$, Google Content Studio$^2$, and Google Research$^3$}

\date{} \abstract{ Dialogue systems for hotel and tourist information
  have typically simplified the richness of the domain, focusing
  system utterances on only a few selected attributes such as price,
  location and type of rooms.  However, much more content is typically
  available for hotels, often as many as 50 distinct instantiated
  attributes for an individual entity. New methods are needed to use
  this content to generate natural dialogues for hotel information, and
  in general for any domain with such rich complex content.  We
  describe three experiments aimed at collecting data that can inform
  an NLG for hotels dialogues, and show, not surprisingly, that 
  the sentences in the original written hotel descriptions provided on
  webpages for each hotel are stylistically not a very good match for
  conversational interaction. We quantify the stylistic features that
  characterize the differences between the original textual data and
  the collected dialogic data. We plan to use these in stylistic
  models for generation, and for scoring retrieved utterances for use
  in hotel dialogues. 
}

\begin{document}
\maketitleabstract

\section{Introduction}
\label{intro-sec}

Research and advanced development labs in both industry and academia
are actively building a new generation of conversational assistants,
to be deployed on mobile devices or on in-home smart speakers, such as
Google Home. None of these conversational assistants
can currently carry on a coherent multi-turn conversation in support
of a complex decision task such as choosing a hotel, where there are
many possible options and the user's choice may involve  making trade-offs among complex
personal preferences and
the pros and cons of different options.  

\begin{figure}[t!]
\begin{center}
\includegraphics[width=3.0in]{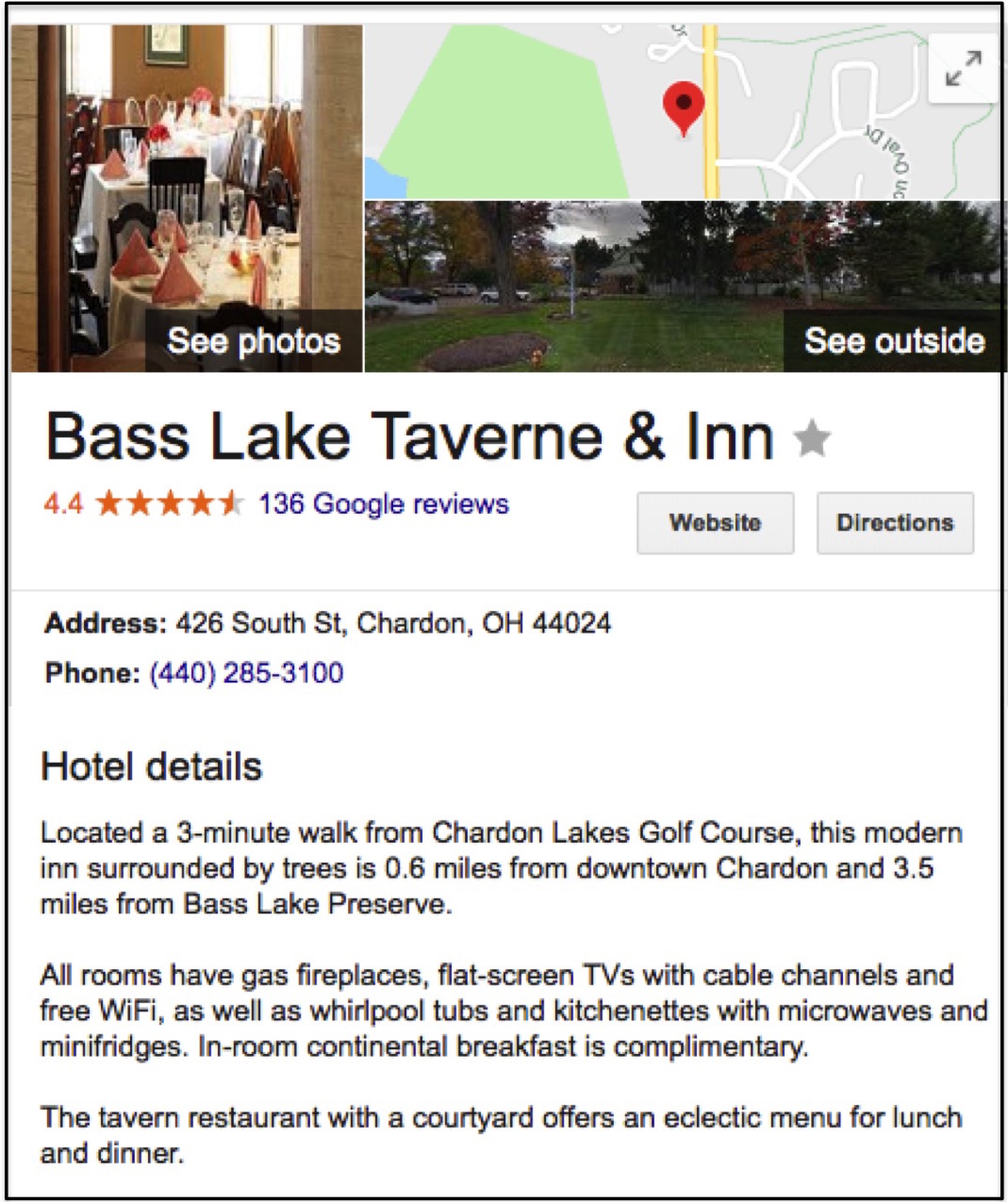}
\vspace{-.1in}
\caption{InfoBox Hotel Description for Bass Lake Taverne \label{CS-desc}}
\end{center}
\vspace{-.3in}
\end{figure}

For example, consider the hotel description in the InfoBox in
Figure~\ref{CS-desc}, the search result for the typed query {\it "Tell
  me about Bass Lake Taverne"}.  These descriptions are written by
human writers within Google Content Studio and cover more than 200
thousand hotels worldwide. The descriptions are designed to provide
travelers with quick, reliable and accurate information that they may
need when making booking decisions, namely a hotel's amenities,
property, and location.  The writers implement many of the decisions
that a dialogue system would have to make: they make decisions about
content selection, content structuring, attribute groupings and the
final realization of the content
\cite{RambowKorelsky92}.  They access multiple
sources of information, such as user reviews and the hotels' own web
pages. The descriptions cannot be longer than 650 characters and are
optimized for visual scanning. There is currently no method for
delivering this content to users via a conversation other than reading
the whole InfoBox aloud, or reading individual sections of it.

\begin{figure}[htb]
\begin{center}
\includegraphics[width=3.2in]{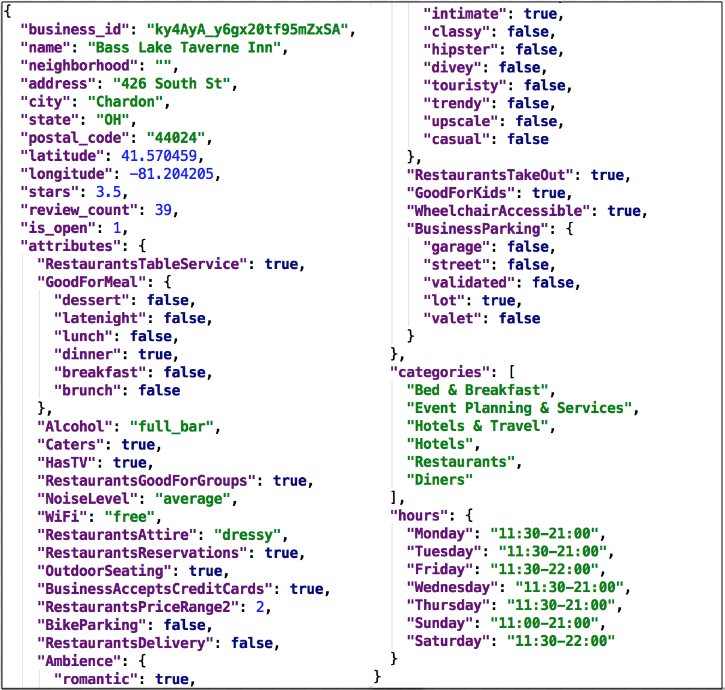}
\caption{Sample of Hotel Structured Data for "Bass Lake Taverne"
 \label{bass-struct}}
\end{center}
\end{figure}

Structured data is also available for each hotel, which includes
information about the setting of a hotel and its grounds, the feel of
the hotel and its rooms,  points of interest nearby, room
features, and  amenities 
such as restaurants and swimming pools. Sample structured data
for the Bass Lake Taverne is  in
Figure~\ref{bass-struct}.\footnote{The publicly available Yelp
  dataset\footnote{\url{https://www.yelp.com/dataset/challenge}} has
  around 8,000 entries for US hotels, providing around 80 unique
  attributes.} The type of information available in the structured
data varies a great deal according to the type of hotel: for
specialized hotels it includes highly distinctive low-frequency
attributes for look-and-feel such as ``feels swanky'' ``historical
rooms'' or amenities such as ``direct access to beach'', ``has hot
tubs'', or ``ski-in, ski-out''.

Research on dialogue systems for hotel information has existed
for many years, in some cases producing shared dialogue corpora that
include  hotel bookings
\cite{devillers2004french,Walkeretal02a,Rudnickyetal99,villaneau2009deeper,bonneau2006results,hastie2002automatic,lemon2006evaluating}.
Historically, these systems have greatly simplified the richness of
the domain, and supported highly restricted versions of the hotel
booking task, by limiting the information that the system can talk
about to a small number of attributes, such as location, number of
stars, room type, and price.  Data collection involved users being
given specific tasks where they simply had to find a hotel in a
particular location, rather than satisfy the complex
preferences that users may have booking hotels.
This reduction in content simplifies the decisions that a dialogue
manager has to make, and it also reduces the complexity of the natural
language generator, since a few pre-constructed templates may suffice
to present the small number of attributes that the system knows about.
It is also important to note that the challenges for the hotel
domain are not unique. Dialogue systems for movies, weather reports,
real estate and restaurant information also have access to rich
content, and yet previous work and current conversational assistants
reduce this content down to just a few attributes.  

This paper takes several steps toward solving the challenging problem
of building a conversational agent that can flexibly deliver richer
content in the hotels domain.\footnote{This version contains updates to the version published at LREC '18 \cite{Walkeretal18}, including updated results.}  Section~\ref{background-overview} first
reviews possible methods that could be applied, and describes several
types of data collection experiments that can inform an initial
design. After motivating these data collection experiments, the rest
of the paper describes them and their results
(Section~\ref{paraphrase-exp}, Section~\ref{generation-exp}, and
Section~\ref{dialogue-exp}).  Our results show, not surprisingly, that
both the hotel utterances and the complete dialogues that we
crowdsource are very different in style than the original written
InfoBox hotel descriptions. We compare different data collection
methods and quantify the stylistic features that characterize their
differences. The resulting corpora are available at {\tt
  nlds.soe.ucsc.edu/hotels}.

\section{Background and Experimental Overview}
\label{background-overview} 

Current methods for supporting dialogue about hotels revolve either
around search or around using a structured dialogue flow. Neither of
these methods on their own support fully natural dialogue, and there
is not yet an architecture for conversational agents that flexibly
combines unstructured information, such as that found in the InfoBox
or in reviews or other textual forms, and structured information such
as that in Figure~\ref{bass-struct}.

Search methods could focus on the content in the current InfoBox, and
carry out short (1-2 turn) conversations by applying compression
techniques on sentences to make them
more conversational \cite{andor2016globally,krause2017redundancy}. For
example, when asked ``Tell me about Bass Lake Taverne'', Google Home
currently produces an utterance providing its location and how far it
is from the user's location. When asked about hotels in a location,
Google Home reads out parts of the information in the Infobox, but it
does not engage in further dialogue that explores individual content
items. Moreover, the
well-known differences between written and oral language
\cite{Biber91} means that selected spans from written descriptions may not sound
natural when spoken in conversation, and techniques may be needed to
adapt the utterance to the dialogic context. Our first experiment,
described in Section~\ref{paraphrase-exp} asks crowdworkers to (1)
indicate which sentences in the InfoBox are most important, and (2)
write  dialogic paraphrases for the selected sentences
in order to explore some of these issues.

Another approach is to train an end-to-end dialogue system for the
hotels domain using a combination of simulation, reinforcement
learning and neural generation methods
\cite{Nayaketal17,shah2018building,liu2017end,gavsic2017spoken}. This
requires first developing a user-system simulation to produce
simulated dialogues, crowdsourcing utterances for each system and user
turn, and then using the resulting data to (1) optimize the dialogue
manager using reinforcement learning,
(2) train the natural language understanding from the user utterances,
and (3) train the natural language generation from the crowd-sourced
system utterances.  Currently however it is not clear how to build a
user-system simulation for the hotels domain that would allow more of
the relevant content to be exchanged, and there are no corpora
available with  example dialogue flows and generated
utterances.

To build a simulation for such complex, rich content, we first need a
model for how the dialogue manager (DM) should (1) order the content
across turns, and (2) select and group the content in each individual
turn.  Our assumption is that the most important information should be
presented earlier in the dialogue, so one way to do this is to apply
methods for inducing a {\it ranking} on the content attributes.
Previous work has developed a model of user preferences to solve this
problem \cite{careninimoore00b}, and shown that users
prefer systems whose dialogue behaviors are based on such customized
content selection and presentation
\cite{Stentetal02,Polifronietal03,Walkeretal07}. These
preferences (ranking on attributes) can be acquired directly from the
user, or can be inferred from their past behavior.  Here we try two
other methods. First, in Section~\ref{paraphrase-exp}, we ask Turkers
to select the most important sentence from the InfoBox descriptions.
We then tabulate which attributes are in the selected sentences, and
use this to induce a ranking. After using this tabulation to collect
additional conversational utterances generated from meaning
representations (Section~\ref{generation-exp}), we carry out an
additional experiment (Section~\ref{dialogue-exp})
where we collect
whole dialogues simulating the exchange of information
between a user and a conversational agent, given particular
attributes to be communicated. We report how information is
ordered and grouped across these dialogues.


An end-to-end training method also needs a corpus for training for the
Natural Language Generator (NLG). Thus we also explore which
crowdsourcing design yields the
best conversational utterance data for training the NLG.  Our first
experiment yields conversationalized paraphrases that match the
information in individual sentences in the original Infobox. Our
second experiment (Section~\ref{generation-exp}) uses content
selection preferences inferred from the paraphrase experiment and
collects utterances generated to match meaning representations.
Our third
experiment (Section~\ref{dialogue-exp}), crowdsources
whole dialogues for selected hotel attributes: the
utterances collected using this method are sensitive to the
context while the other two methods yield utterances that can be used
out of context.

To measure how conversational our collected utterances
are, we build on previous research that
counts linguistic features that vary across different situations of
language use \cite{Biber91}, and tabulates the effect of variables
like the mode of language as well as its setting. We use the
linguistic features tabulated by the Linguistic Inquiry and
Word Count (LIWC) tool \cite{pennebaker2015development}. See
Table~\ref{tab-features}.  We select features to pay attention to
using the counts provided with the LIWC manual that distinguish
natural speech (Column 4) from articles in the New York Times (Column
5). Our hotel descriptions are not an exact genre match to the New
York Times, but they are editorial in nature. For example,
Table~\ref{tab-features} shows that  spoken conversation has
shorter, more common words (Sixltr), more function words, fewer
articles and prepositions, and more affective and social language.

\begin{table}[h!t]
\begin{scriptsize}
\begin{tabular}{llp{.65in}rr} \toprule
Category                     & Abbrev       & Examples        &   Speech & NYT    \\ \toprule
    \multicolumn{5}{l}{\cellcolor[gray]{0.9} \bf Summary Language Variables} \\ 
Words/sentence               & WPS          & -                  &  -  & 21.9    \\
Words \textgreater 6 letters & Sixltr       & -                    &  10.4  &  23.4    \\
    \multicolumn{5}{l}{\cellcolor[gray]{0.9} \bf Linguistic Dimensions}     \\ 
Total function words         & funct        & it, to, no, very     &  56.9  & 42.4    \\
Total pronouns               & pronoun      & I, them, itself      &  20.9  & 7.4    \\
Personal pronouns            & ppron        & I, them, her         &  13.4  & 3.6    \\
1st pers singular            & i            & I, me, mine          &  7.0  &  .6    \\
2nd person                   & you          & you, your, thou      & 4.0   & .3    \\
Impersonal pronouns          & ipron        & it, it's, those      & 7.5   & 3.8    \\
Articles                     & article      & a, an, the           & 4.3   & 9.1    \\
Prepositions                 & prep         & to, with, above      & 10.3   &  14.3    \\
Auxiliary verbs              & auxverb      & am, will, have       & 12.0   &  5.1   \\
Common Adverbs               & adverb       & very, really         &  7.7  & 2.8    \\
Conjunctions                 & conj         & and, but, whereas    &  6.2  &  4.9    \\
Negations                    & negate       & no, not, never       &  2.4  & .6    \\
Common verbs                 & verb         & eat, come, carry     &  21.0  & 10.2    \\
    \multicolumn{5}{l}{\cellcolor[gray]{0.9} \bf Psychological Processes}      \\ 
Affective processes          & affect       & happy, cried         &  6.5  & 3.8    \\
Social processes             & social       & mate, talk, they     &  10.4  &  7.6    \\
Cognitive processes          & cogproc      & cause, know, ought   &  12.3  & 7.5    \\
    \multicolumn{5}{l}{\cellcolor[gray]{0.9} \bf Other}      \\ 
Affiliation                  & affiliation  &  friend, social & 2.0   & 1.7    \\
Present focus                & focuspresent & today, is, now       & 15.3   & 5.1    \\
Informal language            & informal     & -                    &  7.1  &  0.3    \\
Assent                       & assent       & agree, OK, yes       &  3.3  & 0.1    \\
Nonfluencies                 & nonflu       & er, hm, umm          &  2.0  &  0.1    \\
Fillers                      & filler       & Imean, youknow & 0.5 & 0.0\\ \bottomrule
\end{tabular}
\end{scriptsize}
\caption{LIWC Categories with Examples and Differences between Natural Speech and the New York Times \label{tab-features}}
\end{table}

The experiments use Turkers with a high level of qualification 
and we ensure that Turkers make at least minimum wage on our tasks.
For the paraphrase and single-turn HITs for properties and
rooms, we ask for at least 90\% approval rate and at least 100
(sometimes 500) HITs approved and we always do location restriction
(English speaking locations). For the dialog HITs we paid 0.9 per HIT,
and restricted Turkers to those with a 95\% acceptance rate and at
least 1000 HITs approved. We also elicited the dialogs over multiple
rounds and excluded Turkers who had failed to include all 10
attributes on previous HITs.

We present a summary of all of our experiments in
Table~\ref{all-liwc-table} and then discuss the relevant columns in
each section.  A scan of the whole table is highly
informative however, because \newcite{Biber91} makes the point that
differences across language use situations are not dichotomous,
i.e. there is not one kind of oral language and one kind of written
language. Rather language variation occurs continuously and on a
scale, so that language can be ``more or less'' conversational.  The
overall results in Table~\ref{all-liwc-table} demonstrates this scalar
variation, with different methods resulting in more or less
conversationalization of the content in each utterance.

\section{Paraphrase Experiment}
\label{paraphrase-exp}

The overall goal of the Paraphrase experiment is to evaluate the
differences between monologic and dialogic content
that contain the same or similar information.  These experiments are
valuable because the original content is given in unordered lists that
facilitate visual scanning, as opposed to a conversation in which the
dialogue system needs to decide the order in which to present information
and whether to leave some information out.


We ask Turkers to both select "the most important" content out
of the hotel descriptions, and then to paraphrase that content in a
conversational style. We use this data to induce an importance ranking
on content and we also measure how the conversational paraphrases of
that content differ from the original phrasing.  We used a randomly
selected set of 1,000 hotel descriptions from our corpus of 200K, with
instructions to Turkers to:

\begin{small}
\begin{itemize}
\item Select the sentence out of the description that has the most important information
   to provide  in response to a user query to "tell me about HOTEL-NAME".
 \item Cut and paste that sentence into the "Selected Sentence" box.
 \item Rewrite your  selected sentence so that it sounds conversational, as a turn in dialogue.  You may need to reorder the content or convert your selected sentence to multiple sentences in order to make it sound natural.
\end{itemize}
\end{small}

\begin{figure}[ht!]
\begin{center}
\begin{small}
\begin{tabular}{|c|p{2.75in}|}
\hline 
S1 & The elegant rooms, decorated in warm tones, feature high ceilings
and lots of natural light, plus Turkish marble bathrooms, Bose sound
systems, HDTVs and designer toiletries; some have views of the
park. \\
S2 &  Suites include living rooms and soaking tubs; some have city
views.  \\
S3 & Grand suites offer personal butler service. \\
S4& Open since 1930,
this opulent landmark sits across the street from Central Park on New
York's famed 5th Avenue. \\ \hline
\end{tabular}
\end{small}
\end{center}
\caption{\label{another-hotel-desc} An InfoBox description for the hotel {\it The Pierre, A Taj Hotel, New York}, split
into sentences and labeled. } 
\vspace{.1in}
\end{figure}

\begin{figure}[ht!]
\begin{center}
\begin{small}
\begin{tabular}{|c|p{2.65in}|}
\hline 
T1 &  This hotel's elegant rooms are decorated in warm tones. They feature high ceilings with lots of natural light. The rooms feature Turkish marble bathrooms, designer toiletries, high-definition televisions and Bose sound systems. Some rooms even offer views of the park. \\
T2 &  Located on 5th Avenue, this landmark hotel is located across the street from Central Park and dates back to 1930.\\
T3 & Each room is elegantly decorated in warm tones. You will enjoy high ceilings and natural light. The bathrooms are done in Turkish marble and have designer toiletries. For entertainment, you will find HDTVs and Bose sound systems. There are views of Central Park from some rooms. \\
\hline
\end{tabular}
\end{small}
\caption{Turker generated paraphrases of the hotel description shown in Table \ref{another-hotel-desc}. The Turkers T1 and T3 selected S1 as containing the most important information and Turker T2 selected S4. 
\label{another-hotel-utts}} 
\end{center}
\end{figure}

\begin{table}[ht!]
\begin{center}
\begin{scriptsize}
\begin{tabular}{|c|cp{.1in}|}
\hline 
\bf attribute & \bf F&  \\
\hline \hline
locale\_mountain &  1.0 &  \\ \hline
has\_bed\_wall\_in\_rooms &  .67 &  \\ \hline
has\_wet\_room &  .67 &  \\ \hline
feels\_quaint &  .61 &  \\ \hline
has\_crib & .50 &  \\ \hline
feels\_artsy &  .44 &  \\ \hline
is\_whitewashed &  .44 &  \\ \hline
has\_private\_bathroom\_outside\_room &  .44 &  \\ \hline
feels\_nautical &  .42 &  \\ \hline
has\_luxury\_bedding &  .40 &  \\ \hline
welcomes\_children &  .39 &  \\ \hline
is\_dating\_from &  .38 &  \\ \hline
feels\_retro &  .38 &  \\ \hline
all\_inclusive &  .34 &  \\ \hline
has\_casino &  .33 &  \\ \hline
has\_heated\_floor &  .33 &  \\ \hline
has\_city\_views &  .33 &  \\ \hline
has\_boardwalk &  .33 &  \\ \hline
has\_hammocks &  .33 &  \\ \hline
has\_onsite\_barbecue\_area &  .33 &  \\ \hline
\end{tabular}
\end{scriptsize}
\caption{\label{content-selection-counts} Turker's Top 20 Attributes, 
shown with their frequency $F$ of selection when given in the content.}
\end{center}
\end{table}

For each of the descriptions, three Turkers performed this HIT,
yielding a total of 3,000 triples consisting of the original
description, the selected sentence, and the human-generated dialogic
paraphrases. For example, for the hotel description in Figure
~\ref{another-hotel-desc}, two Turkers selected S1 and the other
selected S4. These sentences have different content, so for each
attribute realized we increase its count as part of our goal to induce
a ranking indicating the importance of different attributes.  The
dialogic paraphrases the same Turkers produced are shown in
Figure~\ref{another-hotel-utts}. The paraphrases contain fewer words
per sentence, more use of anaphora, and more use of subjective phrases
taking the listener's perspective such as {\it you will enjoy}.

\begin{table*}[t!hb]
\begin{scriptsize}
\begin{center}
\begin{tabular}{l|rrr||rrcc} \toprule
Category   & InfoBox & Paraphrase & p-val     & Props+Rooms & Dialogues & p-val  & p-val   \\ 
&&              &  &  &  & Props+Rooms vs. Para & Props+Rooms vs. Dial \\ \toprule
Impersonal Pronouns   & 0.97     & 3.80       & 0.00     & 3.36  & 5.19    & 0.00   & 0.00   \\
Adverbs               & 0.97     & 3.41       & 0.00     & 3.57 & 6.25    & 0.12   & 0.00   \\
Affective Processes   & 4.98     & 4.81       & 0.14     & 8.09  & 8.55    & 0.00   & 0.26   \\
Articles              & 8.08     & 9.06       & 0.00     & 11.54 & 7.62    & 0.00    & 0.00   \\
Assent                & 0.02     & 0.04       & 0.00    & 0.07  & 1.13    & 0.11      & 0.00    \\
Auxiliary Verbs       & 1.69     & 6.12       & 0.00    & 8.02  & 11.81   & 0.00       & 0.00    \\
Common Verbs & 3.64     & 7.94       & 0.00      & 10.97 & 15.07   & 0.00       & 0.00   \\
Conjunctions          & 8.07     & 8.13       & 0.54     & 7.33  & 6.52    & 0.00       & 0.00    \\
First Person Singular & 0.01     & 0.02       & 0.00    & 0.41  & 3.41    & 0.00       & 0.00    \\
Negations                  & 0.03     & 0.07       & 0.00       & 0.27  & 0.44    & 0.00  & 0.00      \\
Personal Pronouns     & 0.06     & 1.15       & 0.00    & 3.87  & 10.17   & 0.00       & 0.00   \\

Second Person         & 0.02     & 0.45       & 0.00    & 2.43  & 5.63    & 0.00       & 0.00   \\
Six Letter Words      & 22.21    & 19.15      & 0.00    & 20.74 & 15.50   & 0.00       & 0.00  \\
Social Processes      & 4.81     & 5.63       & 0.00    & 8.53  & 14.66   & 0.00       & 0.00   \\
Total Pronouns               & 1.03     & 4.94       & 0.00     & 7.23  & 15.36   & 0.00       & 0.00    \\
Words Per Sentence       & 22.86    & 14.69      & 0.00     & 14.52 & 10.90   & 0.46       & 0.00   \\
Affiliation                   & 1.18     & 0.95       & 0.00      & 1.13  & 5.97   & 0.08  & 0.00   \\
Cognitive Processes                 & 2.58     & 3.11       & 0.00   & 9.18  & 10.47   & 0.00  & 0.00   \\
Focus present          & 3.64     & 7.91       & 0.00    & 9.35  & 14.40   & 0.00       & 0.00   \\
Function              & 26.79    & 37.62      & 0.00    & 44.58 & 53.08   & 0.00       & 0.00   \\
Informal              & 0.42     & 0.35       & 0.03    & 0.51  & 1.76    & 0.00       & 0.00   \\
nonflu                & 0.37     & 0.28       & 0.00     & 0.42  & 0.63    & 0.00      & 0.00   \\
prep & 9.65     & 8.50       & 0.00    & 9.83 &  9.88    & 0.00       & 0.72    \\

\bottomrule
 \end{tabular}
\end{center}
\end{scriptsize}
\caption{\label{all-liwc-table} Conversational LIWC features across
all Utterance Types/Data Collection Methods}
\vspace{-.2in}
\end{table*}

\noindent{\bf Results.} We build a ranked ordering of hotel attribute
importance using the selected sentences from each hotel
description. We count the number of times each attribute is realized
within a sentence selected as being the most informative or
relevant. We count the number of hotels for which each attribute
applies. The attribute frequency $F$ is given as the number of times
an attribute is selected divided by the product of the number of 
hotels to which the attribute applies and the number of Turkers 
that were shown those hotel descriptions. Finally, the attributes 
are sorted and ranked by largest $F$.

Table~\ref{content-selection-counts} illustrates how the tabulation of
the Turker's selected sentences provides information on the ranking of
attributes that we can use in further experimentation. However, the
frequencies reported are conditioned on the relevant attribute being
available to select in the Infobox description, and many of the
attributes are both low frequency and highly distinctive, e.g.  the
attribute {\tt local\_mountain}. A reliable importance ranking using
this method would need a larger sample than 1000 hotels.
It is also possible that attribute importance should be directly
linked to how distinctive the attribute is, with less frequent
attributes always mentioned earlier in the dialogue.

The first three columns of Table~\ref{all-liwc-table} summarize the
stylistic differences between the original Infobox sentences and the
collected paraphrases. Column 3 provides the p-values showing that
many differences are statistically significant. Differences that
indicate that the paraphrases are more similar to oral language (as in
Speech, column 4 of Table~\ref{tab-features}), include the use of
adverbs, common verbs, and a reduced number of
words per sentence.  Examples of expected differences that are not
realized include increases in affective language and significantly greater use of conjunctions. So while
this method improves the conversational style of the content
realization, we will see that our other methods produce 
{\it more} conversational utterances. While this
method is inexpensive and may not require such expert Turkers, the utterances collected may only
be useful for systems that do not use
structured data and so need paraphrases of the original Infobox
data that is more conversational.

\section{Generation from Meaning Representations}
\label{generation-exp}

The second experiment aims to determine whether we get higher
quality utterances if we ask crowdworkers to generate utterances directly
from a meaning representation, in the context of a conversation,
rather than by selecting from the original Infobox hotel descriptions.
Utterances generated in this way should not be influenced by the
original phrasing and sentence planning in the hotel descriptions.

\begin{figure}[htb]
\begin{center}
\includegraphics[width=3.2in]{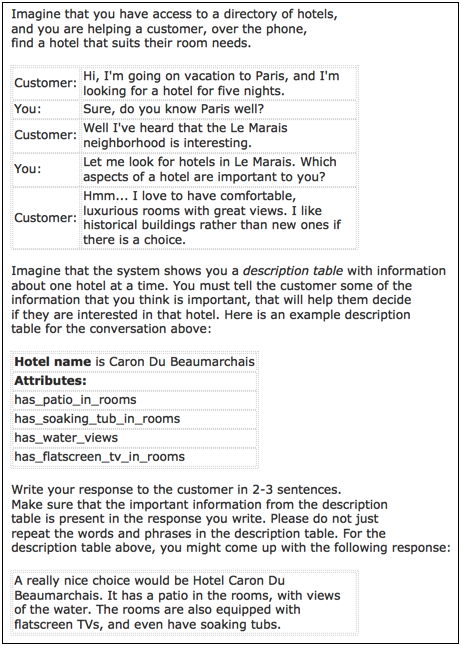}
\vspace{-.2in}
\caption{Instructions for Room Attributes HIT \label{hit-fig}}
\end{center}
\end{figure}

Instructions for our second experiment are shown in
Figure~\ref{hit-fig}.  Here we give Turkers specific content tables
and ask them to generate utterances that realize that content.  Note
that the original hotel descriptions, as illustrated in
Figure~\ref{CS-desc} consists of three blocks of content, property,
rooms and amenities.  For each hotel in a random selection of 200
hotels from the paraphrase experiment, we selected content for both
rooms (4 attributes) and properties (6 attributes) by picking the
attributes with the highest scores (as illustrated for a small set of
attributes in Table~\ref{content-selection-counts}).  Thus hotel has
two unique content tables assigned to it, one pertaining to the
hotel's rooms, and the other for the hotel grounds.  Each hotel
content table is given to three Turkers which results in a total of
1,200 utterances collected.  Turkers were instructed to create
utterances as conversational as possible.

\noindent{\bf Results.} Sample utterances for both properties and
rooms are shown in Figure~\ref{prop-room-utts}.

\begin{figure}[h!bt]
\begin{center}
\begin{small}
\begin{tabular}{|c|p{2.62in}|}
\hline 
Prop &  A good choice is 1 Hotel South Beach in Miami Beach. It's luxurious, lively, upscale, and chic, with beach access and a bar onsite. \\
Prop &  I think that  1 Hotel South Beach will meet your needs. It's a chic luxury hotel with beach access and a bar. Very lively. \\
Room & One of the excellent hotels Miami Beach has to offer is the 1 Hotel South Beach. The upgraded rooms are full featured, including a kitchen and a desk for work. Each room also has a balcony. \\
Room & The 1 Hotel South Beach doesn't mess around. When you come to stay here you won't want to leave. Each upgraded room features a sunny balcony and personal kitchen. You also can expect to find a lovely writing desk for your all correspondence needs. \\
\hline
\end{tabular}
\end{small}
\caption{Example utterances generated by Turkers in the second experiment. Turkers were given specific content tables from which to generate dialogue utterances that realize that content. \label{prop-room-utts}} 
\end{center}
\end{figure}

Column 5 in Table~\ref{all-liwc-table} shows the frequencies of LIWC's
conversationalization features for the utterances collected in this
experiment, and Column 7 reports statistical significance (p-values)
for comparing these collected utterances to the paraphrases collected
in Experiment 1, using an unpaired t-test on the two datasets. We can
see that some of the attributes that indicate conversationalization
indicate that this method yields more conversational utterances: there
are significantly more more auxiliary verbs and common verbs. There is
a greater use of first person and second person pronouns, as well as
words indicating affective, social and cognitive processes.  Counts of
function words and focusing on the present are also higher as would be
expected of more conversational language.

\section{Dialogue Collection Experiment}
\label{dialogue-exp}

\begin{figure}[htb]
\begin{center}
\includegraphics[width=3.2in]{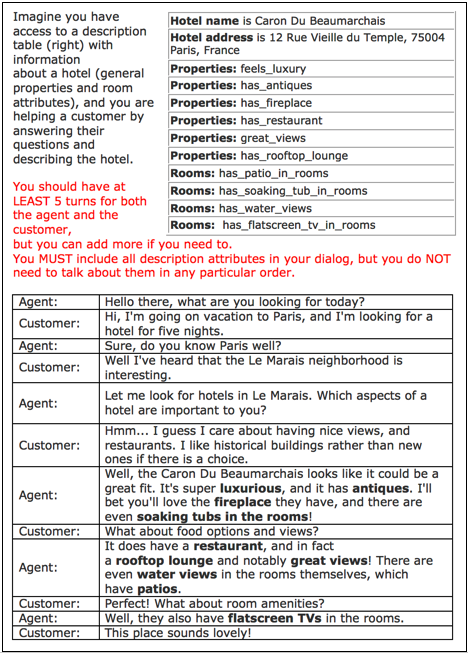}
\vspace{-.2in}
\caption{Instructions for Hotel Dialog HIT\label{hit-fig-dialog}}
\end{center}
\end{figure}

The final data collection experiment focuses on utterance generation
in an explicitly dialogic setting. In order to collect dialogues about
our hotel attributes, we employ a technique called "self-dialogue"
collection, which to our knowledge was pioneered by
\newcite{Krause17}, who claim that the results are surprisingly
natural. We ask individual Turkers to write a full dialogue between
an agent and a customer, where the Turker writes both sides of the
dialogue. The customer is looking for a hotel for a trip, and the
agent has access to a description table with a list of 10 attributes
for a single hotel. The agent is tasked with describing the hotel to
the customer. Figure \ref{hit-fig-dialog} shows our HIT instructions
that provided a sample dialogue as part of the instructions to the
Turkers.

This experiment utilizes 74 unique hotels, a subset of those used in
the property and room experiments above (Section
\ref{generation-exp}), where we have both 6 property attributes, and 4
room attributes. We aimed to collect 3 dialogues per hotel (from
unique Turkers), but due to some Turkers failing to follow
instructions, the final corpus consists of 205 dialogues (comprised of
58 hotels with 3 dialogues each, 15 hotels with 2 dialogs each, and 1
hotel with only 1 dialogue).

Figure~\ref{sit-hotel-dialogues} and Figure
\ref{straight-hotel-dialogues} provide sample dialogues from the
corpus, with the 10 required attributes shown in bold,
and the agent (A) and customer (C) turns shown with their respective
turn numbers. In Dialogue 1, we see an example of a creative dialogue
where the Turker designs a situational context for the dialogue where
the customer is looking for a hotel for a bachelorette party weekend,
and has specific requirements about flooring and amenities. We note
that in this dialogue, the agent only begins to discuss the hotel in
their third turn. In Dialogue 2, we see a much more basic dialogue,
where the agent begins to list properties and room attributes earlier
on the in dialogue (at Agent Turn 2), and the full list of attributes
is exhausted halfway through the conversation, at Agent Turn 3.

\begin{table}[ht!]
\begin{center}
\begin{small}
\begin{tabular}{|c|p{2.7in}|}
\hline 
A1 & Hi! How can I help you today?\\ 
C1 & I am planning a trip to New York for a bachelorette party weekend and need help finding a hotel.\\ 
A2 & OK, what will you girls be planning to do?\\ 
C2 & We're going to a Broadway show but other than that just going to dinner and hitting some bars.\\ 
A3 & OK, I think the Hotel Indigo in the Chelsea section would be great! It's {\bf upscale} and has a great {\bf hip} and {\bf contemporary} vibe with that {\bf buzzy} New York City energy feel.\\ 
C3 & That sounds like what we're looking for. I know this is a weird question but one of the girls sometimes has an allergy to carpet, is there any type of option for a non carpeted room?\\ 
A4 & Actually, this hotel has {\bf hardwood floors in the rooms}.\\ 
C4 & Great! I think we may be bringing some snacks and maybe some of our own alcohol. Can we arrange for a fridge or do they just have ice buckets?\\ 
A5 & The rooms have {\bf mini bars} as well as {\bf coffee} if you girls need some help waking up for your time out. There is also a {\bf bar} on site so you can start the party before you even head out.\\ 
C5 & Great! One more question, one of the girls does need to keep in touch with work. Do you offer WiFi? \\ 
A6 & The hotel has {\bf desks in every room} and offers a {\bf business center} if she needs anything like a printer or a desktop computer. \\ 
C6 & I think we'll go ahead and book this. It sounds perfect!\\ \bottomrule
\end{tabular}
\end{small}
\caption{\label{sit-hotel-dialogues} Situational Context for Content Hotel Dialogue} 
\end{center}
\end{table}

\begin{table}[ht!]
\begin{center}
\begin{small}
\begin{tabular}{|c|p{2.7in}|}
\hline 
A1 & Good evening, how can i help you?\\ 
C1 & I am looking for a good hotel to have a business conference in the Brooklyn area.\\ 
A2 & Sure, let me see what i can find. Hotel Indigo Brooklyn may be just what you are looking for. It has a {\bf hip} feel with an {\bf onsite bar}, {\bf Business center}, {\bf restaurant}, {\bf free wifi}. Its got it all.\\ 
C2 & That sounds excellent. What room amenities are offered?\\ 
A3 & There is {\bf coffee} in the rooms and a {\bf mini fridge}. All the rooms have been recently {\bf upgraded} and did i mention it has a {\bf fitness room}? I has full {\bf room service} as well.\\ 
C3 & Wow, that sounds great.Whats the address? I need to make sure its in the right area for me.\\ 
A4 & Sure, its 229 Duffield Street, Brooklyn, NY 11201, USA.\\ 
C4 & Thanks, thats just the right spot. Go ahead and make me a reservation for next Tuesday.\\ 
A5 & excellent! Its done!\\ 
C5 & Thanks you have been extremely helpful! \\ \bottomrule
\end{tabular}
\end{small}
\caption{\label{straight-hotel-dialogues} Straightforward Attribute Listing Hotel Dialogue} 
\end{center}
\end{table}

\noindent{\bf Results.} We begin by analyzing information that
both the dialogue manager and the natural language generator
would need to know, namely how frequently attributes are
grouped in a single turn in our collected dialogues, by
counting the number of times certain keywords are mentioned related to
the attributes in the dialogues. Table \ref{attribute-grouping} shows
attributes groups that occur at least 4 times in the dataset, showing
the group of attributes and the frequency count. We note that the
attributes within the groups are generally either: 1) semantically
similar, e.g. "modern" and "contemporary"; 2) describe the same
aspect, e.g. "feels elegant" and "feels upscale"; or 3) describe the
same general attribute, e.g. "has breakfast buffet", "has free
breakfast", and "has free breakfast buffet". It is interesting to note
that the semantic similarity is not always completely obvious (for
example, "has balcony in rooms" and "has fireplace" may be used to
emphasize more luxurious amenities that are a rare find).

\begin{table}[ht!]
\begin{center}
\begin{small}
\begin{tabular}{|p{2.5in}|c|} \toprule 
\bf Attribute Group                    & \bf Count \\ \toprule
{ (has\_business\_center, has\_meeting\_rooms) }                                                          & 13 \\ \hline
{ (has\_bar\_onsite, has\_restaurant) }                                                                   & 9  \\\hline
{ (feels\_contemporary, feels\_modern) }                                                                  & 9  \\\hline
{ (has\_bar\_onsite, has\_business\_center) }                                                             & 8  \\\hline
{ (has\_business\_center, has\_desk\_in\_rooms) }                                                         & 7  \\\hline
{ (feels\_casual, feels\_contemporary, feels\_modern) }                                                   & 6  \\\hline
{ (feels\_modern, has\_business\_center) }                                                                & 6  \\\hline
{ (has\_business\_center, has\_convention\_center) }                                                      & 6  \\\hline
{ (feels\_elegant, feels\_upscale) }                                                                      & 4  \\\hline
{ (has\_bar\_onsite, has\_bar\_poolside) }                                                                & 4  \\\hline
{ (has\_microwave\_in\_rooms, has\_minifridge\_in\_rooms) }                                               & 4  \\\hline
{ (feels\_contemporary, feels\_elegant, feels\_modern) }                                                  & 4  \\\hline
{ (feels\_contemporary, feels\_upscale) }                                                                 & 4  \\\hline
{ (has\_balcony\_in\_rooms, has\_fireplace) }                                                             & 4  \\\hline
{ (feels\_chic, feels\_upscale) }                                                                         & 4  \\ \bottomrule
{ (has\_business\_center, \newline has\_desk\_in\_rooms, \newline has\_wi\_fi\_free) }                                      & 4  \\\hline
{ (has\_breakfast\_buffet, \newline has\_free\_breakfast,  \newline has\_free\_breakfast\_buffet) }                          & 4  \\\hline
{ (has\_coffee\_in\_rooms, \newline has\_desk\_in\_rooms, \newline has\_microwave\_in\_rooms, has\_minifridge\_in\_rooms) } & 4  \\\hline
\end{tabular}
\end{small}
\vspace{-.1in}
\caption{\label{attribute-grouping} Attributes Frequently Grouped in a Single Turn}
\end{center}
\end{table}

Our assumption is that more important attributes should be presented
earlier in the dialogue, and that a user-system dialogue simulation
system design \cite{shah2018building,liu2017end,gavsic2017spoken}
would require such information to be available.  Thus, in order to
provide more information on the importance of particular attributes,
we analyze where in the conversation (i.e. first or second half)
certain types of attributes are mentioned. For example, we observe
that attributes describing the "feel", such as "feels chic" or "feels
upscale", are mentioned around 700 times, and that for 80\% of those
times they appear in the first half of the conversation as opposed to
the second half, showing that they are often used as general hotel
descriptors before diving into detailed attributes. Attributes
describing room amenities on the other hand, such as "has kitchen in
rooms" or "has minifridge", were mentioned around 530 times, with a
more even distribution of 53\% in the first half of the conversation,
and 47\% in the second half.

We also observe that most attributes are first
introduced into the conversation by the agent, but that a small number
of attributes are more frequently first introduced by the customer,
specifically: {\it has\_swimming\_pool\_indoor,
  popular\_with\_business\_travelers, has\_onsite\_laundry,
  welcomes\_families, has\_convention\_center, has\_ocean\_view,
  has\_free\_breakfast\_buffet, has\_swimming\_pool\_saltwater}.

Next, we compare our collected dialogues to the single-turn dialogue
descriptions described in Section \ref{generation-exp}). Specifically,
we focus on the "agent" turns of our dialogues, as they are more
directly comparable to the property and room turns.

Table \ref{table-compare-prop-room-dialog} describes the average
number of turns, number of
sentences per turn, words per turn, and attributes per turn across the
property, room, and agent dialogue turns. We note that the average
number of sentences, words, and attributes per turn for our property
and room descriptions are higher in general than the agent turns in
our dialogues, because the dialogues allow the agent to distribute the
required content across multiple turns. 

\begin{table}[ht!]
\begin{center}
\begin{small}
\begin{tabular}{|c|c|c|c|} \toprule 
 &\bf Properties                    & \bf Rooms    & \bf Dialogues       \\ \toprule
\bf Number of turns               & 600      & 600       & 1227      \\
    \multicolumn{4}{l}{\cellcolor[gray]{0.9} \bf Sentences per turn } \\ 
Average                       & 2.80 & 2.55  & 1.80  \\
    \multicolumn{4}{l}{\cellcolor[gray]{0.9} \bf Words per turn } \\ 
Average                       & 41.37 & 39.81 & 21.45\\
    \multicolumn{4}{l}{\cellcolor[gray]{0.9} \bf Attributes per turn } \\ 
Average                       & 6        & 4         & 1.62  \\
\end{tabular}
\end{small}
\caption{\label{table-compare-prop-room-dialog} Comparing Property, Room, and Agent Dialogue Turns}
\end{center}
\end{table}

Column 6 (Dialogues) of Table \ref{all-liwc-table} reports the
frequencies for conversational features in the collected data, with
p-values in Column 8 comparing the dialogic utterances to the
property+room utterances collected in Experiment 2 (Section
~\ref{generation-exp}.  The dialogic data collection results in
utterances that are more conversational according to these counts,
with higher use of impersonal pronouns and adverbs, auxiliary verbs
and common verbs, and first person and second person pronouns.  We
  also see increases in words indicative of affective, social and
  cognitive processes, more informal language, and reduced use of Six 
  Letter words, fewer words per sentence and greater use of language
  focused on the present. Thus these utterances are clearly much more
  conversational, and provide information on attribute ordering across
  turns as well as possible ways of grouping attributes. The
  utterances collected in this way might also be useful for template
  induction, especially if the induced templates could be indexed for
  appropriate use in context.

Finally, Table \ref{table:liwc-examples} presents examples of 
utterances from each dataset for four LIWC categories where we see significant differences across 
the sets, specifically, common verbs, personal pronouns, social processes, and focus present. From the table, 
we can see that for all of these categories, the average LIWC score increases steadily as the data source becomes 
more dialogic.

\begin{table*}[h!]
\begin{scriptsize}
\begin{tabular}
{@{} p{.3in}|p{.3in}|p{.2in}|p{5in} @{}}
\hline
 \bf \scriptsize LIWC Cat. & \bf \scriptsize Dataset & \bf \scriptsize LIWC Avg. & \bf \scriptsize Example \\ \midrule
Common Verbs      & ORIG      & 3.64              & The property has a high-end restaurant and a rooftop bar, as well as 4 outdoor pools, a fitness center and direct beach access with umbrellas.                                                                                                                                                                                     \\\hline
                  & PARA      & 7.94              & A basic hostel that has dorms with four to 8 beds in them. Males and females dorm together in mixed rooms. There is also a roof terrace and pool table.                                                                                                                                                                            \\\hline
                  & PROP+\newline ROOM & 10.97             & Argonaut Hotel is a spectacular historic hotel located just a stones throw from the coast. It has a casual and relaxing yet elegant feel with a touch of  an nautical atmosphere. You can take a stroll along the boardwalk to access Pier 39, or check out Fishermans's Wharf and see all that is surrounding this historic area. \\\hline
                  & DIAL      & 15.07             & Then this hotel would be perfect for you. They have recently updated the place to be more contemporary and have a relaxing touch to it. Also, if you need to meet with your co-workers, they have meeting rooms you can reserve or an on-site restaurant if you need to convene with them.                                         \\\midrule
Personal\newline Pronouns & ORIG      & 0.06              & Straightforward dorm-style \& private rooms with free WiFi, plus a casual bar \& a business center.                                                                                                                                                                                                                                \\\hline
                  & PARA      & 1.15              & The hotel rooms are airy with ocean views and have reclaimed driftwood lining the walls. The rooms come with cotton sheets, live edge wood desks and Nespresso machines. For your entertainment needs, there is free Wi-Fi and 55 inch flat-screen TVs.                                                                            \\\hline
                  & PROP+\newline ROOM & 3.87              & You might want to look at Ocean Resort Fort Lauderdale. It's nautical in theme, but feels modern.  It has an onsite bar, a restaurant, and free wifi.                                                                                                                                                                              \\\hline
                  & DIAL      & 10.17             & Comfort Inn offers roadside lodging so it will be very convenient for you. We also have free parking.                                                                                                                                                                                                                              \\\midrule
Social\newline Processes  & ORIG      & 4.81             & Budget hotel in a converted warehouse The bright, simple rooms come with en suite bathrooms, complimentary Wi-Fi and cable TV. Children age 18 and under stay free with a parent.                                                                                                                                                  \\\hline
                  & PARA      & 5.63              & The rooms in this hotel are nautical-style and have exposed brickwork. Also, they have flat-screen TVs, coffeemakers and yoga mats.                                                                                                                                                                                                \\\hline
                  & PROP+\newline ROOM & 8.53              & The Aston Waikiki Beach Hotel would be the perfect choice. This casual and relaxing hotel offers a host of amenities. It features an onsite, highly-rated restaurant, an outdoor swimming pool, and colorful and fully-outfitted suites.                                                                                           \\\hline
                  & DIAL      & 14.66             & Well, the hotel also has event space for meetings, and there is a bar onsite for winding down with clients afterwards.                                                                                                                                                                                                             \\\midrule
Focus Present     & ORIG      & 3.64              & The polished rooms and suites provide flat-screen TVs, minifridges and Nespresso machines, as well as free Wi-Fi and 24-hour room service                                                                                                                                                                                          \\\hline
                  & PARA      & 7.91              & The hostel offers a relaxed feel and is with-in a few km from several popular destinations. Near by areas include Munich Hauptbahnhof U-Bahn and S-Bahn stations, Oktoberfest, and Marienplatz public square.                                                                                                                      \\\hline
                  & PROP+\newline ROOM & 9.35              & A good choice would be the Ames Boston Hotel, Curio Collection by Hilton. It is historic but has a very modern and chic feeling to it. They offer free wi-fi and a business center for your use.                                                                                                                                   \\\hline
                  & DIAL      & 14.4              & The rooms are relaxing and well appointed. They come with an kitchenette and are equipped with a microwave and a minifridge. Of course there is coffee supplied as well. Are you traveling for business or pleasure?                                        \\ \bottomrule                                                                      
\end{tabular}
\end{scriptsize}
\centering \caption{\label{table:liwc-examples} {Examples for Significantly Different LIWC Categories across the Datasets}}
\end{table*}

\section{Conclusion and Future Work}

This paper presents a new corpus that contributes to defining the
requirements and provide training data for a conversational agent that
can talk about all the rich content available in the hotel domain.
All of the data we collect in all of the experiments is available at
{\tt nlds.soe.ucsc.edu/hotels}.
After completing three different types of data collection, we posit
that the self-dialogue collection might produce the best utterances
but at the highest cost, with the most challenges for direct re-use.
The generation from meaning representations produces fairly high
quality utterances, but they are not sensitive to the context, and
our results from the dialogic collection suggest that it might
be useful to collect additional utterances using this method that
sample different combinations of attributes, and select fewer
attributes for each turn.

In future work, we plan to use these results in three different ways.
First, we can train a "conversational style ranker" based on the data
we collected, so that it can retrieve pre-existing utterances that
have good conversational properties. The features that this ranker
will use are the linguistic features we have identified so far, as
well as new features we plan to develop related to context.  Second,
we will experiment directly using the collected utterances in a
dialogue system, first by templatizing them by removing specific
instantiations of attributes, and then indexing them for their uses in
particular contexts.

\section{References}
\bibliographystyle{lrec}
\bibliography{../../../../nl,../../../../phd}

\end{document}